\documentclass[runningheads]{llncs}
\usepackage[T1]{fontenc}
\usepackage{graphicx}
\usepackage{booktabs}
\usepackage[misc]{ifsym}
\newcommand{\corr}{(\Letter)}
\newcommand\approach[0]{TeMP-TraG}
\usepackage{amsmath} 
\usepackage{amssymb}
\usepackage{caption}
\usepackage{makecell}
\usepackage{multirow}
\usepackage{hyperref}
\DeclareMathAlphabet\mathbfcal{OMS}{cmsy}{b}{n}
\newcommand{\embCal}[1]{\mathbfcal{#1}}
\newcommand{\emb}[1]{\mathbf{#1}}
\usepackage[caption=false]{subfig}

\begin{document}

\title{\approach{}: Edge-based Temporal Message Passing in Transaction Graphs}

\titlerunning{\approach{}: Edge-based Temporal Message Passing in Transaction Graphs}


\author{Steve Gounoue\inst{1,2} \corr \and
Ashutosh Sao\inst{3} \and
Simon Gottschalk\inst{3}}

\authorrunning{S. Gounoue et al.}

\institute{Data Science and Intelligent Systems Group (DSIS), University of Bonn, Bonn, Germany \email{steve.gounoue@cs.uni-bonn.de}
\and
Lamarr Institute for Machine Learning and Artificial Intelligence, Bonn, Germany
\and
L3S Research Center, Hannover, Germany
\email{\{sao,gottschalk\}@l3s.de}}

\maketitle              

\begin{abstract}
Transaction graphs, which represent financial and trade transactions between entities such as bank accounts and companies, can reveal patterns indicative of financial crimes like money laundering and fraud. However, effective detection of such cases requires node and edge classification methods capable of addressing the unique challenges of transaction graphs, including rich edge features, multigraph structures and temporal dynamics. 
To tackle these challenges, we propose \approach{},
a novel graph neural network mechanism that incorporates temporal dynamics into message passing.  
\approach{} prioritises more recent transactions when aggregating node messages, enabling better detection of time-sensitive patterns. We demonstrate that \approach{} improves four state-of-the-art graph neural networks by $6.19\%$ on average. Our results highlight \approach{} as an advancement in leveraging transaction graphs to combat financial crime.

\end{abstract}

\section{Introduction}

Money laundering poses a serious threat to global financial systems, facilitating crimes like fraud, terrorism financing, and corruption. Therefore, there is a need for solutions that can analyse intricate financial transaction graphs and identify the involved actors. 
%
%
%
%
%
%
%
%
%
%
%
Given the inherent graph structure of such networks, Graph Neural Networks (GNNs) are well-suited for this task 
\cite{hall2021efficient} 
through node and edge classification. 
However, they face several challenges:
\begin{enumerate}
    \item Edge features: State-of-the-art GNNs primarily focus on node features during message passing while often disregarding essential edge features, which are critical in transaction graphs. 
    \item Multigraph structures: GNNs typically struggle to effectively model multigraph structures, where multiple edges exist between two nodes -- a common characteristic of financial systems where there are multiple transactions between two nodes. 
    \item Temporal dynamics: The temporal dynamics of transactions, which involve prioritising edges based on their occurrence time, play a crucial role in detecting illicit activities but are often overlooked in prior research.
\end{enumerate}

Recent efforts have addressed these challenges individually: 
%
(i) To incorporate edge features, EGAT merges edge features with node features \cite{DBLP:conf/icann/WangCC21}. However, this approach dilutes edge-specific information, such as relationship types and directional properties, limiting the model’s ability to capture complex relational patterns.
(ii) To handle multigraph structures, Sotiropoulos et al.~\cite{DBLP:conf/bigdataconf/SotiropoulosZLA23} aggregate multiple edges into a single edge using statistical summaries -- potentially overlooking crucial structural details, for example, when an unusually high number of transactions between two nodes signals fraudulent behaviour.
(iii) To model temporal dynamics, T-EGAT~\cite{DBLP:conf/bigdataconf/WangZRGCLD20} and TeMP~\cite{DBLP:conf/emnlp/WuCCH20}\footnote{While TeMP is short for Temporal Message Passing, TeMP actually applies a temporal encoder over embeddings learnt over a sequence of subgraphs split over time, i.e., it does not consider time while performing message passing itself.} split the given graph into multiple subgraphs based on time intervals -- making it difficult to capture meaningful relational patterns over time. Only few approaches such as MEGA-GNN \cite{bilgi2024multigraph} and Multi-GNN \cite{DBLP:conf/aaai/EgressyNBAWA24} tackle more than one of these challenges, while no existing approach simultaneously addresses all three of them, typically lacking comprehension of temporal dynamics.


\begin{figure}[ht]
\centering
\subfloat[Example graph.\label{fig:intro_a}]{\includegraphics[width=0.31\textwidth]{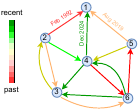}%
}\hfil
\subfloat[Traditional message passing.\label{fig:intro_b}]{\includegraphics[width=0.25\textwidth]{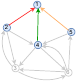}%
}\hfil
\subfloat[Temporal message passing in \approach{}.\label{fig:intro_c}]{\includegraphics[width=0.25\textwidth]{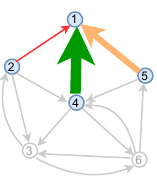}
}

\caption{Comparison of \approach{}'s temporal message passing with traditional message passing, showing that edges are weighted based on temporal proximity.}
\label{fig:intro}

\end{figure}

In this paper, we present \approach{} (\underline{Te}mporal \underline{M}essage \underline{Pa}ssing in \underline{T}rans\-action \underline{G}raphs), a new approach for node and edge representation learning in a transaction graph tackling the three aforementioned challenges: edge features, multigraph structures and temporal dynamics. \approach{} is applied on top of existing graph neural networks such as MEGA-GNN and Multi-GNN, gaining their specific capabilities regarding edge features and multigraph structures comprehension, and further incorporating temporal dynamics through a new temporal message passing paradigm that weights edges based on their temporal proximity.

Fig.~\ref{fig:intro} illustrates how \approach{} captures temporal dynamics: while traditional message passing (Fig.~\ref{fig:intro_b}) aggregates neighbour node messages independent from their temporal characteristics, \approach{} prioritises edges, i.e., transactions, which are more recent to the current node (Fig.~\ref{fig:intro_c}). With this focus on temporal dynamics, \approach{} takes up an important step towards understanding transaction graphs: In financial networks, transaction timing is crucial, as the temporal proximity of events can indicate different behaviours. For example, a burst of transactions within a short period may signal an attempt to rapidly move funds and evade detection, whereas evenly spaced transactions may reflect routine business activity. By weighting transactions based on their temporal proximity, models can prioritise recent interactions that are more indicative of ongoing illicit activities.

Overall, our contribution are as follows:
\begin{itemize}
    \item We propose \approach{}, a novel mechanism that captures temporal dynamics during message passing in a graph neural network.
    \item We demonstrate how \approach{} can be used to extend several existing graph neural networks to make them more time-aware.
    \item Experimental results on two financial transaction graph datasets show that \approach{} improves four state-of-the-art graph neural networks by $6.19\%$ on average. This way, we specifically demonstrate the suitability of our approach towards fighting money laundering and financial crime through the analysis of transaction graphs.
\end{itemize}

\section{Related Work} 

Detecting potentially illicit financial transactions is a crucial task in Anti-Money Laundering (AML). To perform AML, Graph Neural Networks (GNNs) have emerged as a promising methodology. Therefore, in the following, we review (i) graph-based AML and financial fraud detection, (ii) GNNs for multigraph and edge-based learning, and (iii) time-aware GNNs.

\subsection{Graph-Based AML and Financial Fraud Detection}

Graph-based machine learning techniques have become increasingly important in the fight against money laundering and financial fraud. Traditional approaches \cite{frunza2013aftermath,jayasree2016anti,mugarura2011institutional,sarma2020bank} primarily relied on human judgements and expert knowledge to identify and interpret patterns in the data, and also 
often focused on rule-based detection. While these methods provided foundational insights, they struggled to adapt to the dynamic and complex nature of transaction graphs. To address these limitations, recent advances have leveraged machine learning \cite{DBLP:journals/access/ChenSNS21,DBLP:conf/icics/DengRZZ19} and neural networks \cite{DBLP:journals/isafm/GoecksKNSM22,starnini2021smurf,DBLP:journals/symmetry/WanL24} to enhance detection capabilities.

Several graph-based models have been proposed to tackle money laundering detection by leveraging the structural and relational properties of financial transactions. Goecks et al. \cite{DBLP:journals/isafm/GoecksKNSM22} apply GNNs for link prediction and edge classification in temporal transaction graphs, incorporating LightGBM for node classification. This approach successfully models transaction relationships but does not explicitly integrate temporal dependencies into message passing. Similarly, Wan et al. \cite{DBLP:journals/symmetry/WanL24} combine Graph Convolutional Networks with Recurrent Neural Networks to improve pattern detection, capturing sequential transaction patterns but lacking fine-grained control in dynamic networks. Karim et al. \cite{DBLP:journals/access/KarimHCPM24} employ semi-supervised graph learning, utilising topological features for binary classification in AML. However, their model primarily focuses on node-level properties without explicitly refining transaction-level representations.

While these methods highlight the significance of graph-based learning in AML, they often overlook the importance of fine-grained edge-based learning and the dynamics of transaction graphs.

\subsection{GNNs for Multigraphs and Edge-Based Learning}

GNNs have been extensively studied for handling multigraphs and learning edge-based representations to utilise information present on the edges. To this end, edge features have been incorporated into traditional message passing methods such as GIN \cite{DBLP:conf/iclr/XuHLJ19}, which was proved to be a highly expressive GNN architecture and PNA \cite{DBLP:conf/iclr/VelickovicFHLBH19}, which embeds the subgraph structure around a central node of interest into patch representations to enrich graph nodes.

EGAT \cite{DBLP:conf/icann/WangCC21} enriches node features with edge attributes, leveraging an attention-based mechanism to determine the influence of neighbours. Similarly, ADAMM \cite{DBLP:conf/bigdataconf/SotiropoulosZLA23} applies clustering techniques for anomaly detection, utilising the Deep Sets aggregation \cite{zaheer2017deep} for multi-edge representation learning. While effective for multigraphs, EGAT and ADAMM do not fully exploit valuable graph structures due to their aggregations. To exploit such structures, Multi-GNN \cite{DBLP:conf/aaai/EgressyNBAWA24} incorporates ego IDs, bidirectional edges, and port numbering while MEGA-GNN \cite{bilgi2024multigraph} introduces artificial edges to facilitate the bidirectional exchange of information among nodes in the graph. Despite their ability to embed money laundering patterns into node and edge representations, they are not time-aware and fall short in capturing temporal patterns.

Beyond multigraph-specific architectures, edge-based learning techniques such as NNConv \cite{DBLP:conf/icml/GilmerSRVD17} and MGCN \cite{DBLP:conf/aaai/Lu0WHLH19} have been developed for molecular graph classification, encoding edge multiplicity to improve expressivity. However, these models typically assume undirected graphs and predefined edge types, making them less suitable for transaction graphs.

While these approaches contribute significantly to multigraph learning, they do not explicitly incorporate the temporal dimension in message passing.

\subsection{Temporal GNNs and Time-Aware Learning}

AML requires methodologies that are time-aware while preserving structural information of transaction graphs. GNNs need to incorporate time-aware node and edge embedding mechanisms that effectively capture the impact of past transactions on current decision-making. 

One approach towards time-aware learning is to split the transaction graph into multiple graphs valid at different periods. T-EGAT \cite{DBLP:conf/bigdataconf/WangZRGCLD20} performs message passing over these temporal graphs by applying attention-based learning across time steps. Wałęga et al. \cite{DBLP:journals/corr/abs-2408-09918} further distinguish between global and local temporal message passing. 
Tariq et al. \cite{tariq2023topology} instead transform the transaction graph into a temporal graph of sequential transactions. 
While these approaches capture time-dependent relationships to some extent, they do not fully leverage fine-grained transaction timestamps and do not explicitly incorporate edge embeddings, potentially limiting their ability to model transaction-level interactions.

Other works enable time-aware GNNs through recurrent architectures and multi-hop propagation. Wu et al. \cite{DBLP:conf/emnlp/WuCCH20} combine GNNs with Gated Recurrent Units to generate temporal node embeddings. This approach enables sequential dependency learning but does not apply time-aware message passing. Li et al. \cite{DBLP:journals/pvldb/LiSCY23} combine multi-hop message passing with personalised PageRank to make temporal information propagation more efficient.

Although these methods contribute valuable insights into temporal modelling, they typically consider all neighbours of a node equally in message passing, although transactions should have a stronger influence on node and edge embeddings if they are in close temporal proximity. Our work addresses this gap by introducing a time-based weighting mechanism during message passing, ensuring that recent interactions have a greater influence on node and edge embeddings.


\section{Problem Statement}

We aim at detecting potentially illicit activities in transaction graphs, representing actors such as companies and bank accounts and their transactions:

\begin{definition}[Transaction Graph] A transaction graph is defined as $\mathcal{G} = (\mathcal{V}, \mathcal{E}, \mathcal{X}, \mathcal{Z}, \mathcal{T})$,
where: 
\begin{itemize} 
    \item $\mathcal{V}$ is the set of nodes, each representing an entity (e.g., a company, an individual or a bank account).
    
    \item $\mathcal{E}$ is a set of edges, where each edge $(i,j,k) \in \mathcal{E}$ denotes a transaction (e.g., money transfer or item purchase) between entities $i \in \mathcal{V}$ and $j \in \mathcal{V}$. $k \in \{1,2,\dots\}$ indicates the $k$-th edge between $i$ and $j$ as multiple edges can exist between the same pair of nodes, reflecting multiple transactions.
    
    \item $\mathcal{X} \in \mathbb{R}^{|\mathcal{V}| \times d_x}$  is the node feature matrix, where each row corresponds to a node $i \in \mathcal{V}$ and contains a $d_x$-dimensional feature vector (such as creation date or category), denoted as $x_i$. 
    
    \item $\mathcal{Z} \in \mathbb{R}^{|\mathcal{E}| \times d_z}$ is the edge feature matrix, where each row corresponds to an edge $(i,j,k) \in \mathcal{E}$, and contains a $d_z$-dimensional feature vector (such as amount, currency or type), denoted as $z_{i,j,k}$.
    
    \item $\mathcal{T} \in \mathbb{R}^{|\mathcal{E}|}$ is the vector of timestamps associated with each edge, capturing the temporal dynamics of transactions, denoted as $t_{i,j,k}$. 
  
\end{itemize} 
\end{definition}

We denote embedded feature matrices and vectors in bold, i.e., $\embCal{X}$ and $\emb{x}_i$. Further, we define the notion of neighbours as follows:

\begin{definition}[Neighbours]\label{def:neighbours} For a node $i \in \mathcal{V}$ in a transaction graph $\mathcal{G}$, the set of neighbours is defined as:
\begin{align*}
\mathcal{N}(i) = \{ j \in \mathcal{V} \mid (i,j,k) \in \mathcal{E} \vee (j,i,k) \in \mathcal{E} \}
\end{align*}
\end{definition}

Based on these definitions, we define the task of node classification in a transaction graph, for example aiming at classifying entities in a transaction graph as licit or illicit:

\begin{definition}[Node Classification] The node classification task is to learn a function $h: \mathcal{V} \mapsto \mathcal{Y}$
that predicts the label $y_i \in \mathcal{Y}$ for each node $i \in \mathcal{V}$ by leveraging the graph $\mathcal{G}$.
\end{definition}

In analogy, we define the task of \textit{edge classification} $h: \mathcal{E} \mapsto \mathcal{Y}$ that predicts the label $y_{i,j,k} \in \mathcal{Y}$, for instance, to classify specific transactions as illicit or licit.





\section{Approach}

In this section, we introduce \approach{} in detail. We begin with a brief overview, followed by a detailed discussion of \approach{}'s key components, namely, multigraph message passing, temporal message passing, and its training methodology.

\subsection{Overview}

\begin{figure}[t]
\includegraphics[width=\textwidth]{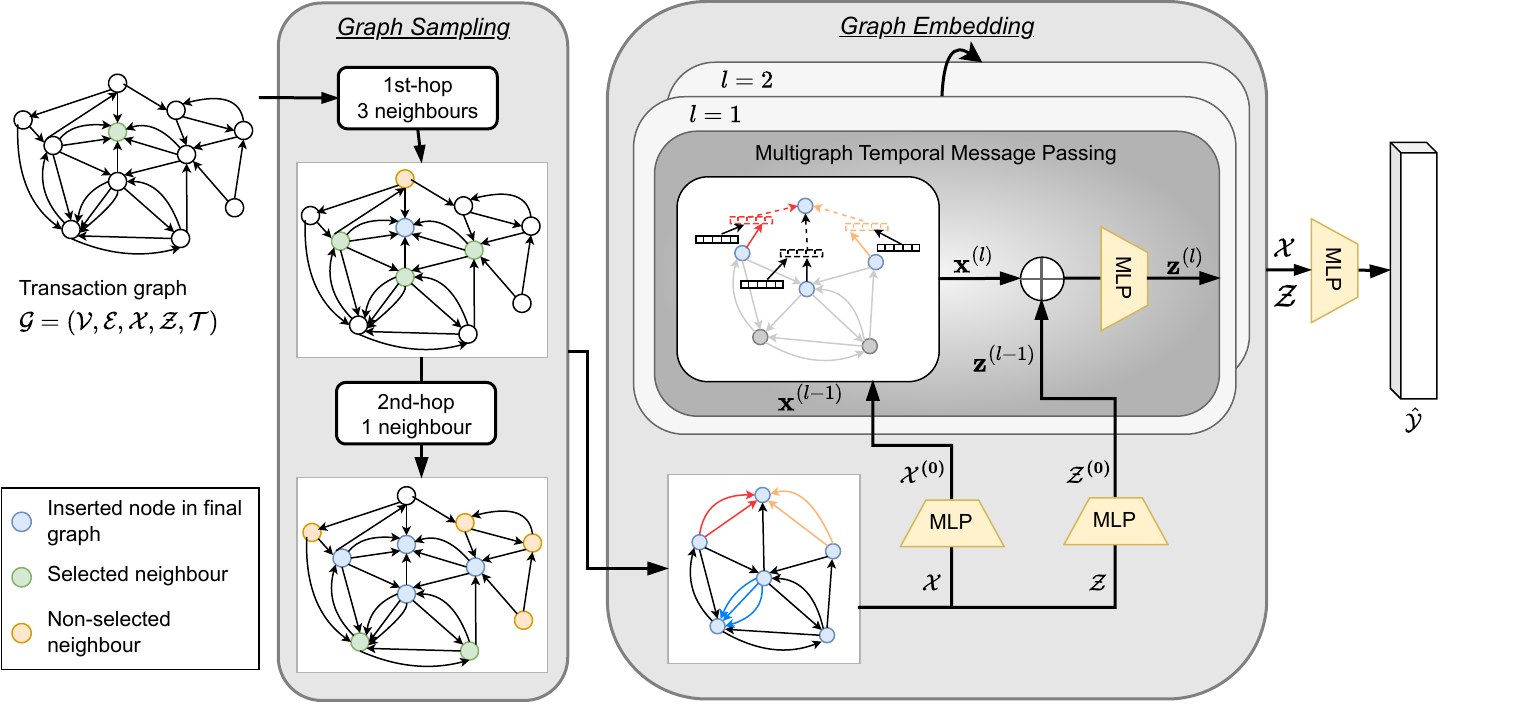}
\caption{Overview of \approach{}. Given a transaction graph $\mathcal{G}$, labels $\hat{\mathcal{Y}}$ are generated for its nodes and edges following graph sampling and embedding.}
\label{fig:pipeline}
\end{figure}

An overview of \approach{} is illustrated in Fig.~\ref{fig:pipeline}. Given the transaction graph $\mathcal{G}$, we first perform graph sampling to deal with the millions of nodes and edges typically contained in transaction graphs. Then, we create an embedding of the graph, i.e., node and edge embeddings, based on our novel multigraph temporal message passing method. Finally, node or edge embeddings are passed through a multilayer perceptron (MLP) to predict the desired labels.

%

\subsubsection{Graph Sampling}
\label{subsubsec:sampling}

To address the computational complexity of transaction graphs, we first sample a subgraph that retains essential structural and relational information while reducing complexity. We employ the GraphSAGE \cite{DBLP:conf/nips/HamiltonYL17} sampling method for efficient learning on large-scale graphs. Specifically, we leverage two-hop neighbourhood sampling, where each node selects a fixed number of its direct neighbours and their neighbours.

\subsubsection{Graph Embedding}

After sampling, \approach{} learns relationships between neighbouring nodes and edges. Since critical information in transaction graphs is embedded in the edges, and multiple edges can exist between two entities, we apply \textit{multigraph temporal message passing}. 
The goal is to learn the representation function $f_{\theta_f}(\cdot)$ that computes latent node representations 
$\embCal{X}$ and edge representations $\embCal{Z}$ by aggregating both node and multi-edge information:
\begin{equation}
    \embCal{X}, \embCal{Z} = f_{\theta_f}(\mathcal{G}),
\end{equation}
where $\embCal{X} \in \mathbb{R}^{|\mathcal{V}| \times \emb{d_x}}$, $\embCal{Z} \in \mathbb{R}^{|\mathcal{E}| \times \emb{d_z}}$ 
and $\theta_f$ are the learnable parameters.


Finally, the prediction component learns a classification function $g_{\theta_g}(\cdot)$ that maps these representations to predicted class labels:
\begin{equation}
    \hat{\mathcal{Y}} = g_{\theta_g} (\embCal{X}, \embCal{Z}),
\end{equation}
where $\theta_g$ are the learnable parameters. 

\subsection{Multigraph Message Passing}
\label{sec:mgmp}

As discussed earlier, edges play a crucial role in transaction graphs, particularly when multiple edges exist between node pairs. To utilise edges, multigraph message passing (MGMP) extends the standard single-graph message passing (SGMP) framework \cite{DBLP:conf/icml/GilmerSRVD17}.

\begin{figure}[t]
    \center

    \subfloat[
    Multigraph.
    \label{fig:edge_aggr_a}]{\includegraphics[width=.3\textwidth]{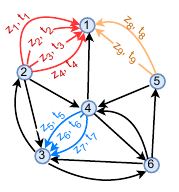}%
    }\hfil
    \subfloat[Aggregation.\label{fig:edge_aggr_b}]{\includegraphics[width=.27\textwidth]{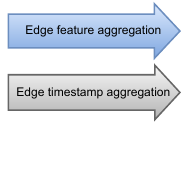}%
    }\hfil
    \subfloat[Aggregated graph.
    \label{fig:edge-aggr_c}]{\includegraphics[width=.3\textwidth]{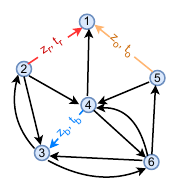}
    }

    \caption{Multigraph transformation by edge feature and timestamp aggregation. For brevity, we label the edges with an indexed edge and timestamp (e.g., $z_1$, $t_1$) and we use colour encoding for aggregated edge (e.g., $z_o$, $t_o$ for orange edges). }
\label{fig:edge-aggr}
\end{figure}

\approach{} employs a MGMP strategy inspired by MEGA-GNN \cite{DBLP:journals/corr/abs-2412-00241}. In addition to \textit{edge feature aggregation}, \approach{} performs \textit{edge timestamp aggregation} 
as illustrated in Fig.~\ref{fig:edge-aggr}. \approach{} simultaneously aggregates edge timestamps into a single representative timestamp leveraging a permutation-invariant function (e.g., max) and aggregates edge features into a single representative feature vector.

Specifically, at layer $l$,  we merge multiple edges with a mean operation:
\begin{equation}
    \emb{z}_{i,j}^{(l)} = \frac{1}{K} \sum_{k=1}^{K} \emb{z}_{i,j,k}^{(l)}, \quad \emb{z}_{i,j,k}^{(0)} = MLP(\emb{z}_{i,j,k}),
\end{equation}
where $K$ is the total number of edges between node $i$ and $j$, and $\emb{z}_{i,j,k}^{(l)}$ represents the embedding of the $k$-th edge between $i$ and $j$.

After edge aggregation, the standard SGMP process is applied, where node embeddings are 
updated through interactions with their local neighbourhoods via \textit{aggregation} and \textit{update} phases.

In the aggregation phase, each node $i$ collects features from its neighbours and the connecting edges from the previous layer $(l-1)$ to construct a message summary $\emb{m}_i^{(l-1)}$ using the aggregation function $\text{AGG}_v(\cdot)$:

\begin{equation}
\label{eq:node-aggr}
    \emb{m}_i^{(l-1)} = \text{AGG}_v \left( \left\{ \left( \emb{x}_j^{(l-1)}, \emb{z}_{i,j}^{(l-1)} \right) \mid j \in \mathcal{N}(i) \right\} \right),
\end{equation}

\noindent where  $\mathcal{N}(i)$ denotes the set of neighbours of node $i$ (see Definition~\ref{def:neighbours}).

In the update phase, the aggregated message $\emb{m}_i^{(l-1)}$ is employed to update the node representation for the layer $(l)$ through the update function $\text{UPD}_v(\cdot)$:
\begin{equation}
\label{eq:node-update}
    \emb{x}_i^{(l)} = \text{UPD}_v\left( \emb{x}_i^{(l-1)}, \emb{m}_i^{(l-1)} \right), \quad \emb{x}_i^{(0)} = MLP(x_i).
\end{equation}

This process is applied simultaneously across all nodes, iteratively updating their embeddings at each layer. 

Finally, utilising the updated embeddings of the neighbour node, the edge embeddings are updated using another update function $\text{UPD}_e(\cdot)$:
\begin{equation}
\label{eq:edge-update}
    \emb{z}_{i,j}^{(l)} = \text{UPD}_e(\emb{x}_i^{(l)}, \emb{x}_j^{(l)}, \emb{z}_{i,j}^{(l-1)})
\end{equation}
By updating edge embeddings alongside nodes, we capture both structural and transactional patterns that influence financial interactions within the graph.

\subsection{Temporal Message Passing}

\begin{figure}
    \centering

    \subfloat[
    Aggregated graph.
    \label{fig:temporal_mp_a}]{\includegraphics[width=.27\textwidth]{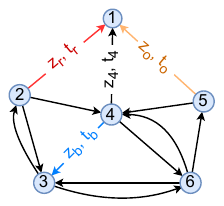}%
    }\hfil
    \subfloat[Update node \textcircled{1}. \label{fig:temporal_mp_b}]{\includegraphics[width=.36\textwidth]{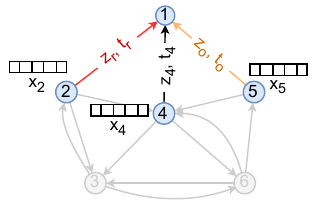}%
    }\hfil
    \subfloat[Weighting messages.
    \label{fig:temporal_mp_d}]{\includegraphics[width=.36\textwidth]{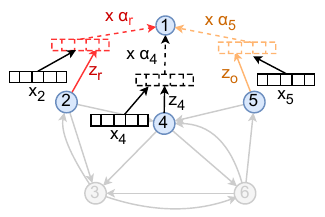}
    }

    \caption{Illustration of temporal message passing with edge aggregation.}
    \label{fig:temporal_mp}
\end{figure}

Temporal information is critical
in detecting money laundering, as illicit financial activities often exhibit temporal dependencies. Recent transactions are especially important for prediction, as they provide the most up-to-date indicators of suspicious behaviour. Unlike traditional message passing, where all neighbours contribute equally, our approach assigns higher importance to more recent neighbours, as illustrated in Fig.~\ref{fig:temporal_mp}.

To achieve time-aware edge prioritisation, we modify the node aggregation function in multigraph message passing, originally defined in Equation \ref{eq:node-aggr}, by incorporating a time-based weighting term:
\begin{equation}
\label{eq:temp-mp-aggr}
     \emb{m}_i^{(l-1)} = \text{AGG}_v \left( \left\{ \alpha_{i,j} \left( \emb{x}_j^{(l-1)}, \emb{z}_{i,j}^{(l-1)} \right) \mid j \in \mathcal{N}(i) \right\} \right),
\end{equation}
where $\alpha_{i,j}$ is a time-based weighting parameter.

\subsubsection{Computation of Time-Based Weighting}

$\alpha_{i,j}$ is computed in two steps:

\begin{enumerate}

    \item \textbf{Effective Timestamp Calculation:} The timestamp of the most recent transaction between two nodes is selected by applying a max-pooling operation:
    \begin{equation}
    \label{eq:time_agg}
        t_{i,j} = \text{max}(t_{i,j,1}, t_{i,j,2}, \dots , t_{i,j,k}).
    \end{equation}

    \item \textbf{Weight Assignment Based on Temporal Importance:} The weighting parameter $\alpha_{i,j}$ is computed by normalising transaction timestamps using a softmax function:
    \begin{equation}
    \label{eq:alpha}
    \alpha_{i,j} = 1 + 
    \frac{exp(t_{i,j})}{\sum_{r \in \mathcal{N}(i)} exp(t_{i,r})}.
    \end{equation}
\end{enumerate}

\subsubsection{\approach{} without Aggregation}


We further propose a configuration of \approach{} to be applied in the SGMP setting, i.e., no edge-featurs aggregation is performed. This way, we can extend GNNs using SGMP (e.g., Multi-GNN). Further, we assume that MGPM is specifically effective in multigraphs with many edges between the same nodes while SGMP could be more effective in sparser graphs. To apply \approach{} in the SGPM setting, Equation \ref{eq:temp-mp-aggr} and Equation \ref{eq:alpha} are replaced as follows:

%

\begin{equation}
    \emb{m}_i^{(l-1)} = \text{AGG}_v \left( \left\{ \alpha_{i,j,k} \left( \emb{x}_j^{(l-1)}, \emb{z}_{i,j,k}^{(l-1)} \right) \mid j \in \mathcal{N}(i), k \geq 1 \right\} \right),
\end{equation}

\begin{equation}
    \alpha_{i,j,k} = 1 + 
    \frac{exp(t_{i,j,k})}{\sum_{r \in \mathcal{N}(i)} \sum_{s \geq 1} exp(t_{i,r,s})}.
\end{equation}


\subsection{Training}

\approach{} is optimised using the cross-entropy loss with L2 regularization to mitigate overfitting:

\begin{equation}
\label{eq:loss}
    \mathcal{L} = -\frac{1}{n} \sum_{i=1}^{n} \sum_{j=1}^{C} y_{ij} \log(\hat{y}_{ij}) + \lambda \sum_{l=1}^{L} \|\mathbf{W}^{(l)}\|_2^2,
\end{equation}
where $\lambda$ is the regularisation strength, $n$ is the number of nodes in node classification (or edges in edge classification), 
$C$ is the number of labels, 
and $\mathbf{W}^{(l)}$ denotes the trainable weights at layer $l$. We use the Adam optimizer with early stopping to prevent overfitting, selecting the best model based on the highest validation F1-score.

\section{Experimental Setup}

In this section, we describe our experimental setup to evaluate \approach{}.


\subsection{Datasets}
%
We evaluate \approach{} on two datasets:  
\begin{itemize}
    \item \textbf{Ethereum Phishing Detection (ETH)} \cite{DBLP:journals/toit/ChenPLLXZ21}: This dataset consists of cryptocurrency transactions from the Ethereum network, where certain accounts are labelled as phishing entities, representing illicit actors.  

    \item  \textbf{IBM Anti-Money Laundering (IBM)} \cite{DBLP:conf/nips/AltmanBNEAA23}: The Small HI (Higher Illicit ratio) dataset is generated using a financial transaction simulator that models interactions between banks, companies, and individuals while incorporating well-established money laundering patterns.
    
\end{itemize}
The dataset statistics are summarised in Table~\ref{tab:dataset}. For both datasets, we follow the preprocessing steps outlined in \cite{DBLP:conf/aaai/EgressyNBAWA24}.

\begin{table}[ht]
     \centering
     \caption{Dataset Statistics. Task: Edge or Node classification.}
     \label{tab:dataset}
     \begin{tabular}{l|rrrrrrrc}
         \toprule
         Dataset & Nodes (N) & Edges (E) & \makecell{Illicit ratio} & \makecell{Licit ratio} & \makecell{Avg. \#edges \\of node pairs} & \makecell{Ratio E/N} & Task \\
         \midrule
         IBM & 30,470 & 5,078,345 & 0.10  \% & 99.90 \% & 17.92 & 166.67 & Edge  \\
         ETH & 2,973,489 & 13,551,303 & 0.04 \% & 99.96 \% & 2.53 & 4.56 & Node  \\
         \bottomrule
     \end{tabular}
 \end{table}

\subsection{Baselines \& Selected Models for Extension}
We compare our approach against multiple baselines spanning feature-engineered machine learning (ML) methods and recent state-of-the-art AML-specific GNNs:

\begin{itemize}
\item \textbf{ML baselines}: We extract graph features from the transaction graph using the Graph Feature Preprocessor by Blanuša et al.~\cite{blanuvsa2024graph} and use them as an input to the following machine learning methods:
\begin{itemize}
    \item \textbf{LightGBM} \cite{DBLP:conf/nips/KeMFWCMYL17}: A gradient boosting decision tree algorithm proposing methods to improve the training efficiency for big data. 
    \item \textbf{Random Forest} \cite{DBLP:conf/icica/LiuWZ12}: A combination 
    of tree structure classifiers to reduce overfitting and improve the robustness against outliers and noise.
    \item \textbf{XGBoost} \cite{DBLP:conf/kdd/ChenG16}: 
    A scalable tree-boosting system leveraging cache access patterns, data compression and sharding.
\end{itemize}
\item \textbf{GNN-based baseline}: Since financial crime detection presents unique challenges, we compare our method against two recent GNN-based AML detection models with two variations each:
\begin{itemize}
    \item \textbf{Multi-GIN} \cite{DBLP:conf/aaai/EgressyNBAWA24} that extends GINs by introducing ego IDs, reverse message passing and port numbering.
    \item \textbf{Multi-PNA} \cite{DBLP:conf/aaai/EgressyNBAWA24}: The same approach based on PNA.
    \item \textbf{MEGA-GIN} \cite{DBLP:journals/corr/abs-2412-00241} that extends GIN with a two-stage aggregation process in the message passing layer, first parallel edge aggregation, followed by a node-level aggregation of messages from distinct neighbours.
    \item \textbf{MEGA-PNA} \cite{DBLP:journals/corr/abs-2412-00241}:  The same approach based on PNA.
\end{itemize}

\end{itemize}

To evaluate how \approach{} improves existing GNN models, we extend GIN \cite{DBLP:conf/iclr/XuHLJ19} and PNA \cite{DBLP:conf/iclr/VelickovicFHLBH19}.

\subsection{Metrics}
We employ three metrics during evaluation:
\begin{itemize}
    \item \textbf{F1-min}: The F1-score of the minority class to measure the model’s ability to recall illicit transactions while maintaining precision; 
    \item \textbf{Macro F1}: The average of the F1-scores across all classes, providing a balanced assessment of both illicit and licit classifications; 
    \item \textbf{PR-AUC}: Precision-Recall Area Under the Curve to assess the robustness of the model with an unknown decision boundary. 
\end{itemize}

\section{Results}
\label{sec:results}

We evaluate \approach{} following the experimental setup described before. First, we compare \approach{} to the baselines. Then, we examine the different aggregation strategies for edge timestamps during the message passing. Finally, we explore the impact of graph sampling by varying the number of selected first- and second-hop neighbours.


\subsection{Overall Results}

Table \ref{tab:results} presents the overall performance comparison of different models across multiple datasets and evaluation metrics. In few cases, we observe that traditional feature-engineered ML methods achieve results comparable to AML-specific GNN models (e.g., XGBoost outperforms Multi-GIN). 
However, in most of the cases, the GNN-based approaches outperform the ML-based approaches by a considerable margin.
Our approach \approach{} consistently enhances the different GNN model architectures it extends. For example, \approach{} has a mean absolute improvement over the GNN models of $14.72\%$ and $1.96\%$ regarding the Macro F1 on ETH and IBM, respectively, making \approach{} the best-performing method in our evaluation.

\begin{table}[ht]
    \centering
    \caption{Performance comparison of different models across datasets (in \%). We highlight the best results in \textbf{bold} and \underline{underline} the second-best results. The last row presents the average absolute improvement of \approach{} over the GNN models. \textit{with and w/o agg} indicate whether we perform edge feature aggregation during message passing for node representation learning or not.
    }
    \begin{tabular}{|l|rrr|rrr|}
        \hline
        & \multicolumn{3}{c|}{\textbf{ETH}} & \multicolumn{3}{c|}{\textbf{IBM}} \\
        \cline{2-7} 
        \textbf{Model} & F1-min & Macro F1 & PR-AUC & F1-min & Macro F1 & PR-AUC \\
        \hline \hline
        LightGBM & 35.27 & 67.62 & 32.49 & 48.97 & 74.45 & 50.36 \\
        Random Forest & 24.39 & 62.18 & 44.80 & 42.45 & 71.19 & 61.34 \\
        XGBoost & 48.21 & 74.10 & 56.34 & 62.13 & 81.04 & 68.17 \\
        \hline
        Multi-GIN & 32.48 & 66.20 & 49.73 & 60.07 & 80.00 & 43.69 \\
        Multi-PNA & 62.83 & 81.40 & 56.39 & 67.11 & 83.53 & 60.58 \\
        MEGA-GIN & 54.87 & 77.42 & 49.28 & 70.38 & 85.16 & 67.09 \\
        MEGA-PNA & 47.49 & 73.74 & 48.76 & \underline{71.52} & 85.74 & \underline{68.88} \\
        \hline \hline
        \approach{} (GIN w/o agg) & 63.29 & 81.64 & 56.35 & 61.36 & 80.64 & 61.44 \\
        \approach{} (PNA w/o agg) & \textbf{66.20} & \textbf{83.09} & \underline{57.58} & 68.52 & 83.73 & 61.55 \\
        \approach{} (GIN with agg) & \underline{64.66} & \underline{82.32} & \textbf{59.05} & 71.14 & \underline{86.04} & 68.08 \\
        \approach{} (PNA with agg) & 62.48 & 81.23 & 55.38 & \textbf{75.92} & \textbf{87.94} & \textbf{73.86} \\
        \hline \hline
        Mean absolute improvement & $+$14.72 & $+$7.37 & $+$5.93 & $+$1.96 & $+$0.98 & $+$6.17 \\
        \hline
    \end{tabular}
    \label{tab:results}
\end{table}

For the ETH dataset, \approach{} (PNA w/o agg) outperforms the other models on F1-min with a $1.54\%$ improvement over the second-best and $14.72\%$ improvement on average. For the IBM dataset, \approach{} (PNA with agg) achieves the best performance for all metrics with an improvement of $1.96\%$ on average regarding F1-min.

Further, we observe that \approach{} yields stronger improvements on the ETH dataset when applied without aggregation (w/o agg) during message passing, whereas, for the IBM dataset, it is most effective with aggregation (with agg).
This discrepancy can be attributed to the structural differences: ETH has considerably more edges per node pair than IBM (see Table~\ref{tab:dataset}). Edge aggregation becomes crucial in handling dense edge structures, while its benefits diminish when fewer edges exist between a pair of nodes.
%

Overall, our results confirm that \approach{} outperforms all baselines across both datasets and all evaluation metrics. 

\subsection{Analysis of Edge Timestamp Aggregation Methods}

\begin{figure}[ht]
\centering
\subfloat[F1-min.\label{fig:aggr_f1_minority}]{\includegraphics[width=0.33\textwidth]{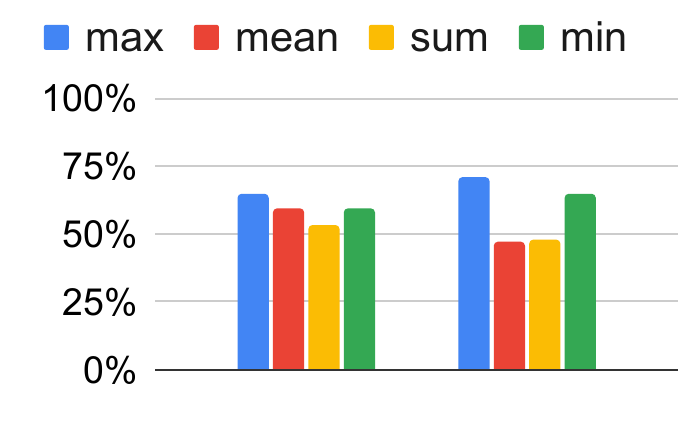}%
}\hfil
\subfloat[Macro F1.\label{fig:aggr_macro_f1}]{\includegraphics[width=0.33\textwidth]{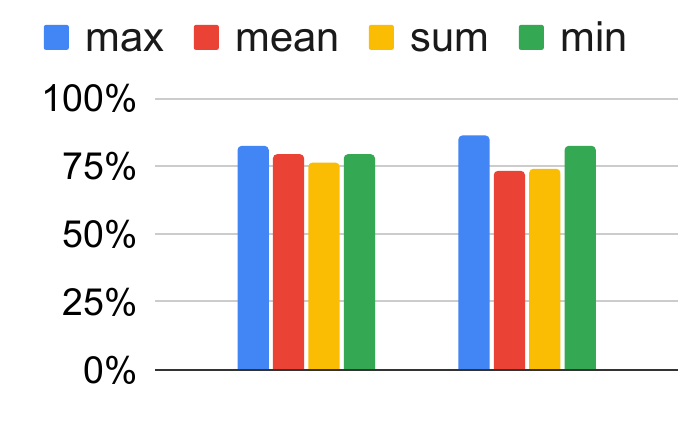}%
}\hfil
\subfloat[PR-AUC.\label{fig:aggr_pr_auc}]{\includegraphics[width=0.33\textwidth]{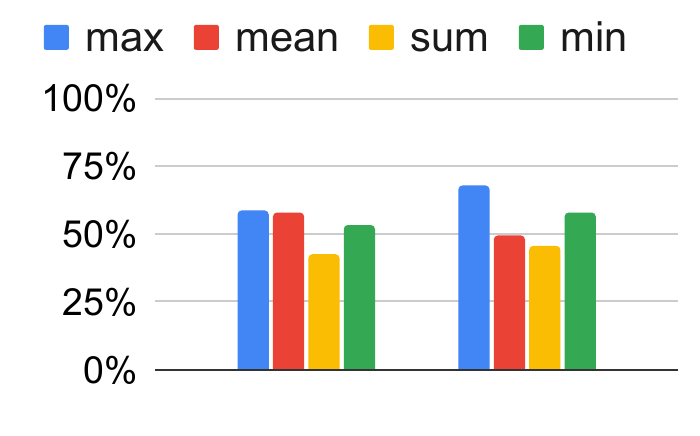}
}

\caption{Performance of different aggregation strategies for edge timestamps in \approach{} (GIN with agg) according to three metrics.}
\label{fig:sensitivity_aggregation}

\end{figure}

Fig.~\ref{fig:sensitivity_aggregation} presents the performance of \approach{} on both datasets using four permutation-invariant aggregation strategies for edge timestamps: sum, mean, min, and max. Each strategy prioritises a different aspect of temporal information: \textit{sum} emphasises transaction volume, \textit{mean} balances between older and recent transactions,
\textit{min} gives precedence to older transactions, and \textit{max} favours more recent interactions (see Equation~\ref{eq:time_agg}).

\textit{sum} performs the worst in four out of six cases; in the other two cases, \textit{mean} performs the worst. Both aggregations encapsulate all  transactions, making it impossible to derive findings about a specific single transaction between two nodes.
In contrast, \textit{min} and \textit{max} maintain strong performance across all metrics and datasets, with \textit{max} achieving the best overall results.

%

\subsection{Impact of Graph Sampling}

\begin{figure}[ht]
\centering
\subfloat[F1-min (ETH).\label{fig:sampling_f1_minority_eth}]{\includegraphics[width=0.33\textwidth]{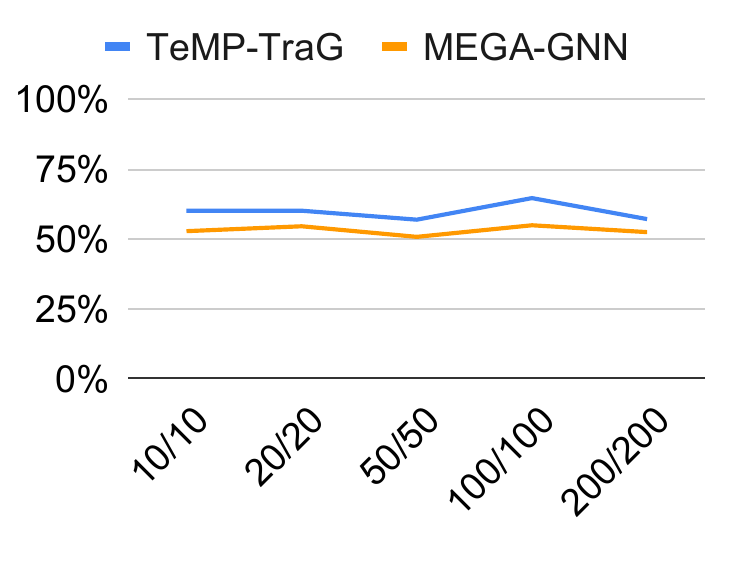}%
}\hfil
\subfloat[Macro F1 (ETH).\label{fig:sampling_macro_f1_eth}]{\includegraphics[width=0.33\textwidth]{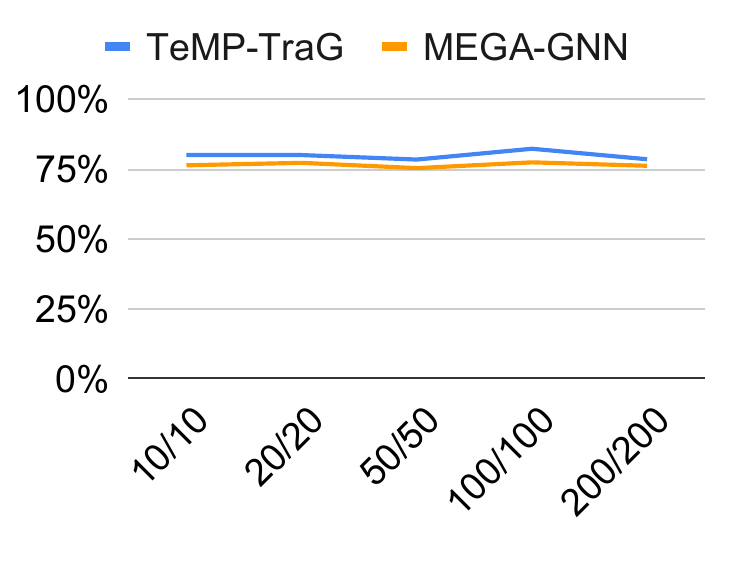}%
}\hfil
\subfloat[PR-AUC (ETH).\label{fig:sampling_pr_auc_eth}]{\includegraphics[width=0.33\textwidth]{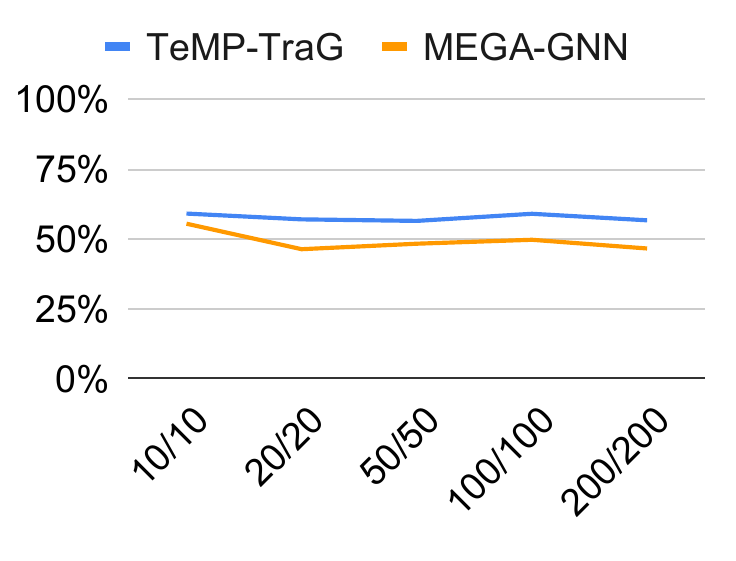}
}\hfil
\subfloat[F1-min (IBM).\label{fig:sampling_f1_minority_ibm}]{\includegraphics[width=0.33\textwidth]{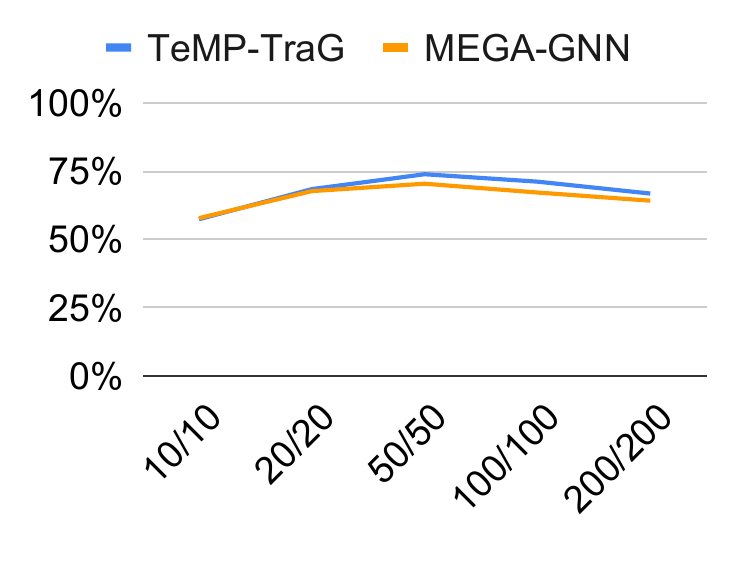}%
}\hfil
\subfloat[Macro F1 (IBM).\label{fig:sampling_macro_f1_ibm}]{\includegraphics[width=0.33\textwidth]{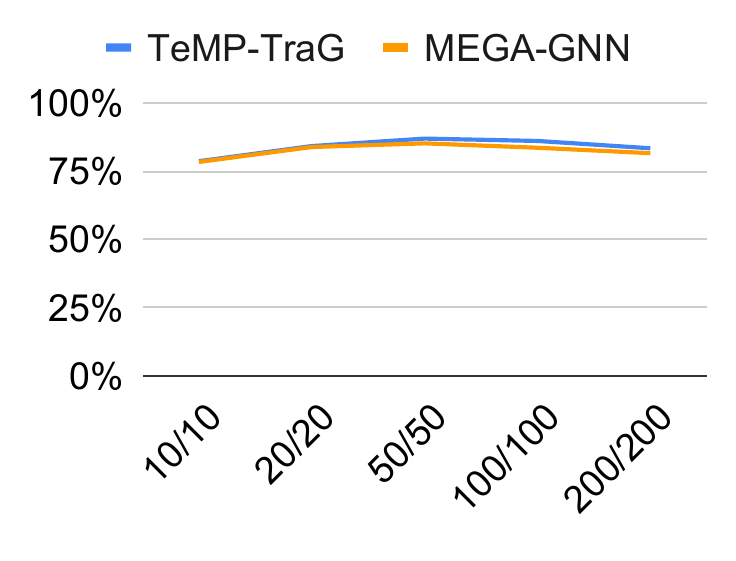}%
}\hfil
\subfloat[PR-AUC (IBM).\label{fig:sampling_pr_auc_ibm}]{\includegraphics[width=0.33\textwidth]{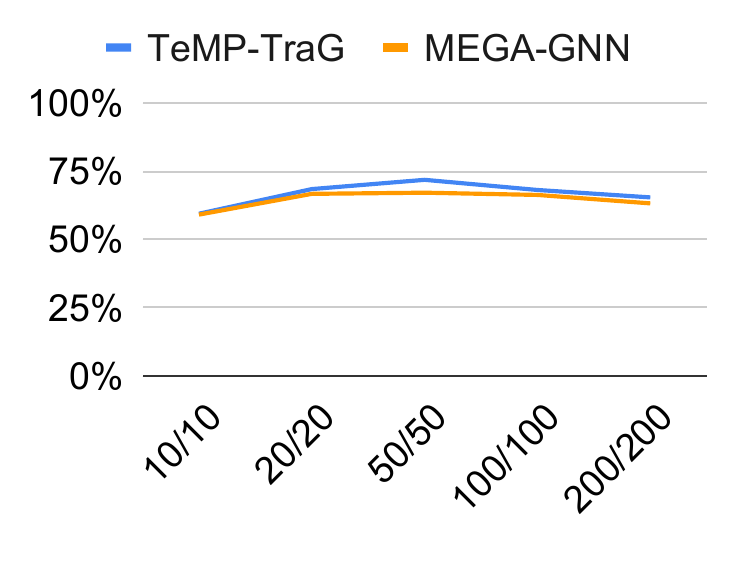}
}

\caption{Impact of graph sampling in \approach{} (GIN with agg) compared to MEGA-GIN. The x-axes indicate the number of node neighbours for the first- and second-hop during graph sampling. 
}
\label{fig:graph_sampling}

\end{figure}

Next, we 
study how the number of first- and second-hop neighbours during graph sampling (Section~\ref{subsubsec:sampling}) affects model performance. Fig.~\ref{fig:graph_sampling} illustrates the performance evolution of \approach{} (GIN with agg) and MEGA-GIN 
as the neighbourhood size increases.

For ETH, we observe minimal performance variation as the number of sampled neighbours increases, i.e., robust behaviour. 
Conversely, for IBM, the performance initially improves with more neighbours before stabilising: For very few neighbours, the performance gap between MEGA-GNN and \approach{} is narrow (lower than $0.5\%$ for all metrics) and then increases up to $2\%$ with more than $50/50$ 1-hop/2-hop neighbours. 

Overall, \approach{} demonstrates consistent benefits across various graph sampling settings, maintaining competitive performance with state-of-the-art models even when only a small number of neighbours is selected.

\section{Conclusion \& Future Works}

In this work, we introduced \approach{} -- a novel graph neural network mechanism that incorporates temporal dynamics into message passing to address the core challenges in transaction graphs.
\approach{} effectively handles edge features for multigraph embedding, incorporating a temporal weighting mechanism in the message-passing neural network. By prioritising recent transactions for graph embedding, \approach{} enhances the ability of GNNs to capture time-sensitive patterns, leading to more effective detection of suspicious activities in financial transaction graphs. 
\approach{} improves four state-of-the-art graph neural networks by $6.19\%$ on average.

There is potential to further advance graph learning in transaction graphs: First, incorporating geographical information will enable to capture spatio-tem\-po\-ral patterns in financial crime. 
Second, GNNs for AML could heavily benefit from the domain knowledge of AML experts by incorporating their knowledge, e.g., rules reflecting common illicit activity patterns.


\bibliographystyle{splncs04}
\bibliography{references}

\end{document}